\documentclass[11pt]{article}
\usepackage{hyperref}
\usepackage{emnlp2017}
\usepackage{times}
\usepackage{latexsym}
\usepackage{amsmath}
\usepackage{amssymb}
\usepackage{amsfonts}
\usepackage{bbm}
\usepackage{bm}
\usepackage{multirow}
\usepackage{graphicx}
\usepackage{color}
\usepackage{sidecap}
\usepackage{enumitem}
\usepackage{booktabs}

\emnlpfinalcopy

\def\emnlppaperid{1095}

\DeclareMathOperator*{\argmax}{arg\,max}

\newcommand{\draftonly}[1]{#1}
\renewcommand{\draftonly}[1]{}

\newcommand{\ra}[1]{\renewcommand{\arraystretch}{#1}}

\newcommand{\draftcomment}[3]{\draftonly{\textcolor{#2}{{\textbf{[#3 --\textsc{#1}]}}}}}
\newcommand{\sw}[1]{\draftcomment{ss}{teal}{#1}}
\newcommand{\nascomment}[1]{\draftcomment{nas}{red}{#1}}
\newcommand{\sam}[1]{\draftcomment{sam}{blue}{#1}}

\newcommand{\todo}[1]{\draftonly{\textcolor{red}{{\textbf{[\textsc{todo}:
#1]}}}}}

\newcommand{\interalia}[1]{\citep[\textit{inter alia}]{#1}}

\newcommand{\fnframe}[1]{\textsc{#1}}
\newcommand{\fnrole}[1]{\textsc{#1}}
\newcommand{\term}[1]{\emph{#1}} 
\newcommand{\1}[1]{\mathbbm{1}_{#1}}
\newcommand{\tensor}[1]{\mathbf{#1}}
\newcommand{\seq}[1]{\bm{#1}}
\newcommand{\tuple}[1]{\langle #1 \rangle}

\newcommand{\sect}[1]{(\S\ref{#1})}

\newcommand{\rolesForF}{\mathcal{Y}_f}
\newcommand{\nullLabel}{\fnrole{null}}

\newcommand{\hidden}{\tensor{h}}
\newcommand{\bilstm}{\overrightarrow{\overleftarrow{\tensor{biLSTM}}}}
\newcommand{\lstm}{\overrightarrow{{\tensor{LSTM}}}}

\newcommand{\relu}{\tensor{reLU}}
\newcommand{\segScore}{\phi}

\usepackage{xspace}
\newcommand{\sysname}{Open-SESAME}

\setitemize{noitemsep,topsep=6pt,parsep=0pt,partopsep=0pt,itemindent=0pt,leftmargin=12pt,labelwidth=12pt}
\setenumerate{noitemsep,topsep=6pt,parsep=0pt,partopsep=0pt,itemindent=0pt,leftmargin=12pt,labelwidth=12pt}

\title{Frame-Semantic Parsing with Softmax-Margin Segmental  RNNs \\and a Syntactic Scaffold}

\author{Swabha Swayamdipta$^{\clubsuit}$ ~ Sam Thomson$^{\clubsuit}$ ~ Chris Dyer$^{\diamondsuit}$ ~ Noah A. Smith$^{\heartsuit}$\\
  $^\clubsuit$School of Computer Science, Carnegie Mellon University,
  Pittsburgh, PA, USA\\
  $^\diamondsuit$Google DeepMind, London, UK \\
$^\heartsuit$Paul G.~Allen School of Computer Science \& Engineering, University of
Washington, Seattle, WA, USA
\\
 {\tt \{swabha,sthomson,cdyer\}@cs.cmu.edu, nasmith@cs.washington.edu} \\ 
  }
\date{}

\begin{document}
\maketitle

\begin{abstract}
We present a new, efficient frame-semantic parser that labels semantic
arguments to FrameNet predicates.
Built using an extension to the segmental RNN that emphasizes recall, our basic system
achieves competitive performance without any calls to a syntactic parser.
We then introduce a method that uses phrase-syntactic annotations from the Penn Treebank during training only,
through a multitask objective;
no parsing is required at training or test time.
This ``syntactic scaffold'' offers a cheaper alternative to traditional
syntactic pipelining, and achieves state-of-the-art performance.

\end{abstract}


\nascomment{also, please make font size in figure 3 larger (ideally
  close to the caption font size}
\section{Introduction}
\label{sec:intro}

\begin{figure*}
\begin{center}
  \includegraphics[width=1.6\columnwidth]{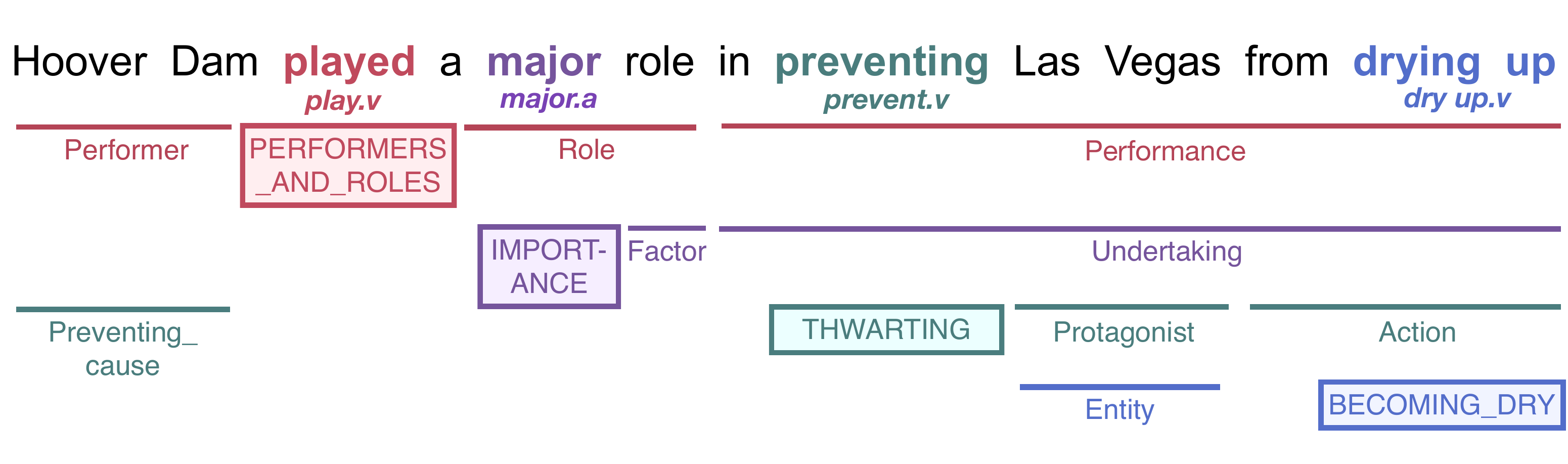}
\end{center}
  \caption{
    A FrameNet sentence with color-coded frame annotations below.
    Target words and phrases are highlighted in the sentence, and their lexical units are shown italicized below.
    Frames are shown in colored blocks, and frame element segments are shown horizontally alongside the frame. 
  \label{fig:dryingup}}
\end{figure*}

Frame-semantic parsing~\citep{Gildea:02a} is the task of identifying the semantic frames evoked in text, along with their arguments, 
formalized in the FrameNet project~\citep{Baker:98}.
An example sentence and its frame-semantic annotations are shown in
Figure~\ref{fig:dryingup}.
Frame-semantics has shown to be useful in question answering \cite{Shen:07}, text-to-scene generation~\cite{Coyne:12}, dialog systems \cite{Chen:13} and social-network extraction \cite{Agarwal:14}, among others.

Existing frame-semantic parsers rely on syntactic parsing in two ways: first, syntax-based rules are used to prune out spans of text that are unlikely to contain any frame's argument; and second, syntax provides features for argument identification and labeling \cite{Das:14,Fitzgerald:15,Tackstrom:15,Roth:15}.
Syntactic parsing comes at a computational cost, though, and syntactic filters are known to be too strict.
Indeed, \citet{Tackstrom:15} found that filtering heuristics based on predicted dependencies bounded recall below 72.6\%.
Recent research has begun to question whether syntax is necessary for semantic analysis \citep{Zhou:15,Swayamdipta:16,He:17,Peng:17}, and here, we begin by posing the same question for frame semantics in particular.

Our first model presents the first syntax-free frame-semantic argument identification system~\sect{sect:argid},
which achieves competitive results.
We follow \citet{Kong:16} in combining bidirectional RNNs with a semi-Markov CRF---a model known as the segmental RNN (\term{SegRNN})---to segment and label the sentence relative to each frame.
We introduce the \term{softmax-margin SegRNN}~\sect{sect:softmaxmargin}, a modification to the model that we use to encourage recall over precision.
This approach to argument identification forgoes syntactic filtering and syntactic features completely.\footnote{
We also release a syntax-free frame identification system~\sect{sect:frameid}. However, the focus of this paper is argument id. For controlled comparison, we use the results from a state-of-the art frame id system~\cite{Hermann:14}, which uses syntactic features.} 

In the rest of the paper, using this basic model as a foundation, we test whether incorporating syntax is still worthwhile.
In the first setting~\sect{sect:synfeats}, we add standard features derived from either a dependency parser or a phrase-structure parser to the softmax-margin SegRNN.
We find that this syntactic pipelining approach improves over both our syntax-free model and achieves state-of-the-art performance.
This finding confirms that syntactic features are beneficial, even on top of expressive neural representation learning.

In our second setting~\sect{sect:scaffold}, we preserve most of the benefit of syntactic features without the accompanying computational cost of feature extraction or syntactic parsing.
We incorporate the training objective of our syntax-free model into a multitask setting where the second task is unlabeled constituent identification (i.e., a separate binary decision for each span).
This task is trained on the Penn Treebank, sharing the underlying sentence representation with the frame-semantic parser.
This \term{syntactic scaffold}\footnote{We borrow the term \term{scaffolding} from developmental psychology \citep{Wood:76} to describe a support
task during learning that is eventually discarded.} task offers useful guidance
to the frame-semantic model, leading to performance on par with our models that use syntactic features.
This approach also achieves state-of-the-art performance, despite not involving a syntactic parser during training or testing.

To summarize, our contributions are:
\begin{enumerate}
  \item[a.] the softmax-margin SegRNN, a recall-oriented extension to segmental RNNs, for frame-semantic argument identification without any syntactic information~\sect{sect:argid},
  \item[b.] the addition of syntactic information to the above, achieving state-of-the-art perfomance, using:
  \begin{enumerate}
    \item[i.] a pipelined approach, incorporating features from automatic dependency or phrase-structure parsers~\sect{sect:synfeats},
    \item[ii.] a syntactic scaffolding approach, discarding the need for a syntactic parser altogether~\sect{sect:scaffold}.
  \end{enumerate}
\end{enumerate}
Our open-source implementation is available as \term{open-SESAME} (SEmi-markov Softmax-margin ArguMEnt parser) at \url{https://github.com/Noahs-ARK/open-sesame/}.

\section{Frame-Semantic Parsing Task}
\label{sect:task}

The Berkeley FrameNet project \cite{Baker:98,Ruppenhofer:10} provides
a lexicon of 1,020 semantic frames,\footnote{Release 1.5. Release 1.7
  has 1,222 frame annotations; see \url{http://framenet.icsi.berkeley.edu}.} a corpus of
sentences annotated with frames from that inventory, and a corpus
of annotated exemplar sentences (not used in this work).

Each \term{frame} represents a kind of event, situation, or relationship, and
has a set of \term{frame elements} (semantic roles) associated with
it~\cite{Fillmore:76}.
In a sentence, frames are evoked by \term{targets}, which are words or phrases.
The FrameNet lexicon maintains a list of \term{lexical units} for each frame, which are lemma and part-of-speech pairs that can evoke that frame.
For example, in Figure~\ref{fig:dryingup}, the target \textit{drying up}
has \textit{dry up.v} as its lexical unit, associated with the frame \fnframe{Becoming\_dry}.
Our main use of the FrameNet lexicon, following earlier work, is as a mapping
between frames and the roles they might take.

Frame-semantic parsing is usually performed as a pipeline of tasks:
target identification (which words or expressions evoke frames?), frame
identification (which frame does each target evoke?), and then argument
identification (for each frame $f$, and each of its possible roles in
the FrameNet-defined set $\rolesForF$, which
span of text provides the argument?).
Target identification conventionally  relies on
heuristics, and frame identification is usually treated as a
classification problem \cite{Das:14}. 
The focus of
this paper is argument identification, and we evaluate variations of
our approach on both gold-standard frame input and on the output of 
state-of-the-art frame identification~\citep{Fitzgerald:15}\todo{add reference to our own frame id again?}.

\subsection{Formal Notation}
\label{sect:notation}

A single input instance for argument identification consists of:  an $n$-word sentence $\seq{w}_d = \tuple{w_{d_1}, w_{d_2}, \ldots,
  w_{d_n}}$, its predicted part-of-speech tag sequence, $\seq{w}_o= \tuple{w_{o_1}, w_{o_2}, \ldots,
  w_{o_n}}$, a single target span $t = \tuple{t_{\text{start}}, t_{\text{end}}}$,
  its lexical unit $\ell$, and its evoked frame
  $f$. 
For brevity we denote the input as  $\seq{x}= \tuple{\seq{w}_d, \seq{w}_o, t, \ell, f}$.
Given this input $\seq{x}$, the task is to produce a \term{segmentation} of the
sentence: $\seq{s}=\tuple{s_1, s_2, \ldots, s_m}$. 
The $k$th \term{segment} $s_k = \tuple{i_k, j_k, y_k}$ corresponds to  a labeled 
\term{span} of the sentence,  with start index $i_k$, end index $j_k$, and label $y_k$.
The label $y_k \in \rolesForF \cup \{ \nullLabel \}$ is either the role that the
span fills, or \nullLabel\ if the segment does not fill any role.
The segmentation is constrained so that argument spans cover the
sentence and do not overlap ($i_{k+1} = j_k + 1$; $i_1 = 1$; $j_m = n$); segments of length $1$ such that $i_k = j_k$ are allowed.
A separate segmentation is produced for each frame evoked in
a sentence.

\section{Segmental RNN for Argument Identification}
\label{sect:argid}

Our first model for argument identification is a segmental RNN
\citep[SegRNN;][]{Kong:16}, a variant of a semi-Markov conditional
random field~\citep{Sarawagi:04} whose span representations are computed using
bidirectional RNNs.
Semi-Markov CRFs model a conditional distribution over
labeled segmentations of an input sequence; precisely the set of
outputs possible for a single frame's argument identification task.
They provide $O(n^2)$ inference using dynamic programming~\sect{sect:softmaxmargin}.
This can be reduced to $O(nb)$ by filtering out segments longer than $b$ tokens (we use $b=20$, which prunes less than 1\% of gold arguments).

Semi-Markov models are more general than BIO tagging schemes, which have been
used successfully in PropBank SRL~\interalia{Collobert:11,Zhou:15}.
The semi-Markov assumption allows scoring functions that directly model an
entire variable-length segment (rather than fixed-length label n-grams), while
retaining exact inference and a linear runtime.
Relatedly, \citet{Tackstrom:15} introduced a dynamic program that allows
direct modeling of variable-length segments as well as enforcing constraints such as
certain roles being filled at most once.
Its runtime is linear in the sentence length, but exponential in the number of roles.
The semi-Markov CRF's inference algorithm is a relaxed special case of their method, with fewer constraints and without the exponential runtime constant.

SegRNNs use continuous vector representations of spans.  In past work,
they have been applied to joint word segmentation and part-of-speech
tagging for Chinese \citep{Kong:16} and to speech recognition \citep{Lu:16}.

\begin{figure}[th]
\centering
  \includegraphics[width=0.95\columnwidth]{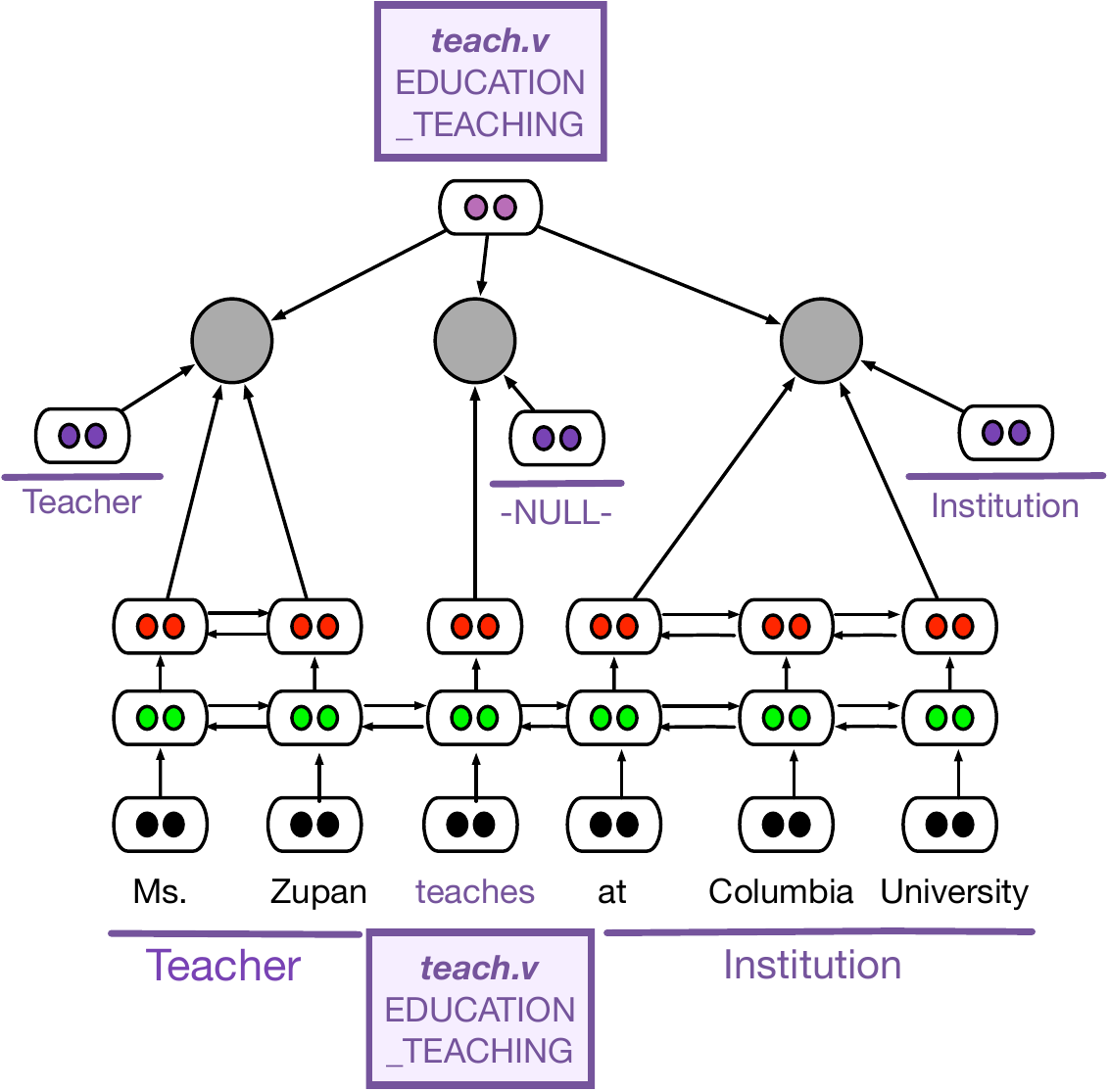}
  \caption{ Illustration of the model architecture for an example sentence and its frame-semantic parse. 
  The input token embeddings are depicted in black, and the input frame and frame-element embeddings in purple.
  The token biLSTM hidden states are shown in green, and the span embedding hidden states in red.
  A final multi-layer perceptron connects all the different components
  of the model into a segment factor, shown in gray.
This analysis contains three segments, one for each of the two frame
elements and a third \nullLabel\ segment. 
  \label{fig:model}}
\end{figure}

Given an input $\seq{x}$, a SegRNN  defines a conditional distribution $p(\seq{s} \mid \seq{x})$.
Every segment $s_i$ is given a real-valued score $\segScore(s_i,
\seq{x})$, detailed in \S\ref{sect:segScore}.
The score of a segmentation $\seq{s}$ is the sum of the scores of its segments:
\begin{align}
\segScore(\seq{s}, \seq{x}) = &
\displaystyle\sum_{i=1}^{m}
  \segScore(s_i, \seq{x}).
\end{align}
These scores are exponentiated and normalized to define the
probability distribution.
The sum-product variant of the semi-Markov dynamic programming
algorithm is used to calculate the normalization term (required during
learning).  At test time, the max-product variant returns the most probable segmentation,
\begin{align}
\hat{\seq{s}} =& \argmax_{\seq{s}}
\segScore(\seq{s}, \seq{x}).
\end{align}
The parameters of SegRNNs are learned  to maximize a criterion related
to the conditional
log-likelihood of the gold-standard segments in the training corpus
\sect{sect:softmaxmargin}.
The learner evaluates and adjusts segment scores $\segScore(s_i, \seq{x})$~\sect{sect:segScore} for every span in the sentence,
which in turn involves learning embedded representations for all
spans~\sect{sect:spanemb}.  Representations of the target, lexical
unit, frame, and frame elements  are also learned~\sect{sect:tfemb}. 
The entire model is illustrated in Fig.~\ref{fig:model}; we discuss
the details from the bottom up.

\subsection{Input Span Embeddings}
\label{sect:spanemb}

We use two  bidirectional long short-term memory
networks~(\citealp[biLSTMs;][]{Hochreiter:97,Schuster:97,Graves:12})
over the input sentence to obtain continuous representations of each token and
each candidate span (all subsequences of length $\le 20$).

At each word position $q$, we give as input to the first (token) biLSTM an input vector
$\tensor{v}_q = [\tensor{d}_{q}; \tensor{e}_{q}; \tensor{o}_q; \boldsymbol{\gamma}_q]$, where
$\tensor{d}_{q}$ is a learned embedding of the word type, $\tensor{e}_{q}$
is a fixed pre-trained embedding of the word type, $\tensor{o}_{q}$ is a
learned embedding of the part-of-speech tag, and $\boldsymbol{\gamma}_q$ is the distance of the word from the beginning of the target.
This yields a hidden state vector $\hidden^{\text{tok}}_q$ at the $q$th
token, which is a contextualized representation of the token:
\begin{align}
\hidden^{\text{tok}}_q =&
  \left[\bilstm^{\text{tok}}(
    \tensor{v}_1,
    \ldots,
    \tensor{v}_n)
  \right]_q.
\end{align}

The second biLSTM embeds every candidate span, using the hidden representations from the token biLSTM as input:
\begin{align}
\hidden^{\text{span}}_{i:j} =&
  \bilstm^{\text{span}}(
    \hidden^{\text{tok}}_{i},
    \ldots,
    \hidden^{\text{tok}}_j
  ).
\end{align}
\sam{We need to explain that we use the last token state from the forward LSTM concatenated with the first token state from the backward LSTM.}
These span embeddings are calculated efficiently, sharing computation
where possible; for details, see \citet{Kong:16}.

\subsection{Input Target and Frame Embeddings}
\label{sect:tfemb}

In addition to the token and span embeddings above, we learn an embedding $\tensor{v}_f$ for each frame $f$, and an embedding
$\tensor{v}_\ell$ for each lexical unit $\ell$.
To represent the target in context, we use $\hidden^{\text{tok}}$ over the target
span $t$, as well as the neighboring token on each side, as an input to a forward LSTM:
\begin{equation}
\tensor{v}_t =
  \lstm^{\text{tgt}}(
    \hidden^{\text{tok}}_{t_{\text{start}}-1},
    \ldots,
    \hidden^{\text{tok}}_{t_{\text{end}}+1})
  .
\end{equation}
The above are concatenated to form a representation of the
target and frame, which is used in representing segments~\sect{sect:segScore}: $\tensor{v}_{f, \ell, t} = [\tensor{v}_f; \tensor{v}_\ell; \tensor{v}_t ]$

\subsection{Segment Scores}
\label{sect:segScore}

The score of a segment should capture the interaction between the span, the frame
element label, the target, the lexical unit, and the frame.
We form a vector $\tensor{v}_s$ for a segment $s=\tuple{i,j,y}$ by concatenating the span embedding $\hidden^{\text{span}}_{i:j}$,
a learned embedding $\tensor{v}_y$ for frame element $y \in \rolesForF \cup \{\nullLabel\}$,
and two additional one-hot features (denoted $\boldsymbol{\mu}$):
the binned length of the span, and the span's position relative to the
target (before, after, overlapping, or within):
\begin{align}
\tensor{v}_s =&
  [
    \hidden^{\text{span}}_{i:j} ;
    \tensor{v}_y; \boldsymbol{\mu}
  ].
\label{span-rep}
\end{align}
Then the representation is passed through a rectified linear
unit~\cite{Nair:10} to get the segment score:
\begin{align}
\label{eqn:segScore}
\segScore(s,\seq{x}) =&
  \tensor{w}_2 \cdot \relu \{
    \tensor{W}_1 [\tensor{v}_{s}; \tensor{v}_{f,\ell,t}]
  \},
\end{align}
where the matrix $\tensor{W}_1$ and the vector $\tensor{w}_2$ are
model parameters.
Note that \nullLabel-labeled spans are handled in
the same way as other labeled spans.\footnote{
Our formalization allows spurious ambiguity, in that the spaces in
between arguments may segmented into any number of \nullLabel\ segments.
We experimented with requiring \nullLabel\ segments to have length $1$
to eliminate the ambiguity, but found that performance dropped
significantly.}

\subsection{Softmax-Margin Segmental RNNs}
\label{sect:softmaxmargin}
Most spans are not arguments;
we therefore train to maximize a criterion that biases the
model to favor recall \citep{Mohit:12}.  Known as the softmax-margin \citep{Gimpel:10},
this criterion alters the partition function with a cost function that
more strongly penalizes false negatives:
\begin{align}
\label{eqn:softmax-margin-segrnn}
\text{loss}(\seq{x},\seq{s}^*) =
  - \log \frac{\exp \segScore(\seq{s}^*,\seq{x})}{Z}, \\
  Z =
    \displaystyle\sum_{\seq{s}}
    \exp{ \{ \segScore(\seq{s}, \seq{x})  +
      \text{cost}(\seq{s},\seq{s}^*) \}
    }, \\
  \text{cost}(\seq{s},\seq{s}^*) =
    \alpha \text{FN}(\seq{s},\seq{s}^*) +
    \text{FP}(\seq{s},\seq{s}^*),
\end{align}
where $\text{FN}$ counts false negatives, $\text{FP}$ counts false positives,
and $\alpha \geqslant 1$ is a hyperparameter tuned on the development set.
In order to keep inference tractable, the cost function needs to factorize by predicted span (so $\text{cost}(\seq{s}, \seq{s}^*) = \sum_{i=1}^m { \text{cost}(s_i, \seq{s}^*) }$).
But a false negative is not a property of an individual predicted span.
We get around this by noting that a predicted span forces a false negative if it
partially overlaps with a gold span.
To avoid double counting, as multiple predicted spans may overlap with a gold span,
we assign blame only to the span that contains the first token ($i^*$) of the missing gold span:
$\text{cost}(\tuple{i, j, y}, \seq{s}^*) =$
\begin{align}
  \begin{cases}
  0,
      & \text{ if } \tuple{i, j, y} \in \seq{s}^* \\
  \1{y \neq \nullLabel} +
    \displaystyle \hspace{-1em}\sum_{\substack{
      \tuple{i^*, j^*, y^* \neq \nullLabel} \in \seq{s}^* \\
      i \leqslant i^* \leqslant j
    }} \hspace{-1em}  \alpha ,
      & \text{ otherwise. }
  \end{cases}
\end{align}

The softmax-margin criterion, like log-likelihood, normalizes globally
over all of the exponentially many possible labeled segmentations.
The following zeroth-order semi-Markov dynamic program \citep{Sarawagi:04} efficiently computes $Z$:
\begin{align}
z_j =& 
  \displaystyle\sum_{
    \substack{
      s = \tuple{i, j, y} \\
      j - i < 20
    }
  } {
    z_{i-1} \exp \{ \segScore(s, \seq{x}) + \text{cost}(s, \seq{s^*}) \}
  } 
\end{align}
where $Z = z_n$, under the base case $z_0 = 1$.  The cost
function is easily incorporated because it factors in the same way as
$\segScore$.

The model's prediction can be calculated using a similar dynamic
programming algorithm with the following recurrence (and the usual
``arg max'' bookkeeping):
\begin{align}
b_j =
  \displaystyle\max_{\substack{
    s = \tuple{i, j, y} \\
    j - i < 20
  }} {
    b_{i-1} \exp \segScore(s, \seq{x}),
  }
\label{eq:dp}
\end{align}

Our model formulation enforces only the no-overlap constraint; we expect it learn other SRL constraints from the data.
We optimize using ADAM~\citep{Kingma:14}.
Models are trained a single thread on an NVIDIA Tesla K40 CPU.\sam{If we don't have timing numbers, I don't think we need for this detail.}
Convergence requires $15$ epochs.
Hyperparameters are tuned based on performance on the held-out development set, with further implementation details given in~\S\ref{sect:experiments}.

\section{Syntactic Features}
\label{sect:synfeats}

Syntax has been important in many past models for semantic argument
prediction~\interalia{Punyakanok:08,Toutanova:08,Johansson:08,Fitzgerald:15}.
The SegRNN-based model in \S\ref{sect:argid} is syntax-free, but it is
straightforward to incorporate features from a syntactic parse of
the sentence as  additional input to the model.  We note that
the computational cost of syntactic parsing---especially
phrase-structure parsing---is significant, but it is important
to test whether the SegRNN model can benefit from syntax.

\paragraph{Phrase-structure features.}  We apply a state-of-the-art
phrase-structure parser \citep[RNNG;][]{Dyer:16}, extracting the features in Table~\ref{tab:phrasefeats} for each span.
We then concatenate these features to the span representation (Equation~\ref{span-rep})
in our span scoring model.
We found that 84\% of gold-standard argument spans correspond to
predicted constituents.

\begin{table}
\begin{footnotesize}
\begin{centering}
\ra{1.2}
\begin{tabular*}{.9\columnwidth}{@{}p{1.8cm} p{4.7cm}@{}}
  \toprule
  Name & Description \\
  \midrule
  \texttt{is\_phrase} &
    fires if the span is any constituent (binary)
    \\
  \texttt{phrase\_type} &
    type of constituent, or \nullLabel\ (one-hot)
    \\
  \texttt{lca\_type} &
    the phrase type of the least common ancestor of the target and the span\
    (one-hot)
    \\
  \texttt{path\_lstm} &
    final state of an biLSTM over the sequence of phrase types on the path from
    the target to the span\ (dense vector)    \\
  \bottomrule
\end{tabular*}
\caption{Phrase-structure syntactic features.
\label{tab:phrasefeats}}
\end{centering}
\end{footnotesize}
\end{table}

\paragraph{Dependency features.} We apply a state-of-the-art dependency parser
\citep[SyntaxNet;][]{Andor:16}, extracting the features in
Table~\ref{tab:depfeats} from its output.
The two word-level features are concatenated to the word vectors before they
are fed into the token BiLSTM.
The three span-level features are concatenated to the span representation
(Equation~\ref{span-rep}).
The \texttt{out\#} features capture information similar to that used in prior
work's span-filtering heuristics.
The \texttt{out\#=1} feature, for example, is an approximation of the
\texttt{is\_phrase} constituent feature.
Since word-level representations include dependency head and label
information, the \texttt{path\_lstm} has access to that information for
each dependency along the path as well.

Finally, note that these two variants of our model do not filter
candidate spans, as done in past work.  Instead, we allow the learner
to flexibly consider syntactic
features when scoring spans, potentially leading to a kind of ``soft'' filter.

\begin{table}
\begin{footnotesize}
\begin{centering}
\ra{1.2}
\begin{tabular*}{.9\columnwidth}{@{}p{1.8cm} p{4.7cm}@{}}
  \toprule
  Name & Description \\
  \midrule
  \textit{token-level:} \\
  \texttt{head\_word} &
    the word vector of the head of this token (dense vector)
    \\
  \texttt{head\_label} &
    the head dependency label (one-hot)
    \\
    \\
  \textit{span-level:} \\
  \texttt{out\#} &
    binned value of the number of tokens in the span which have heads outside the span
    (one-hot)
    \\
  \texttt{path\_lstm} &
    biLSTM over the sequence of token representations on the
    dependency path from the target to the span\ (dense vector), similar to \citet{Roth:16}
    
    \\
  \bottomrule
\end{tabular*}
\caption{Dependency syntax features.
\label{tab:depfeats}}
\end{centering}
\end{footnotesize}
\end{table}

\section{Syntactic Scaffolding}
\label{sect:scaffold}

A wide range of recent results in NLP have shown the benefit of multitask representation learning~\cite{Caruana:97}, where the same embeddings of words are learned to minimize the loss functions of multiple tasks~\cite{Luong:15,Chen:17,Peng:17}.   
Here we consider a ``scaffold'' task---one we use only during learning, and
whose output we are not especially interested in---in a multitask setup with our
basic model~\sect{sect:argid}.
The second task in our setup is learning to predict syntactic constituent
spans.
Since frame-semantic arguments are often also constituents, we hypothesize that
learning which text spans are constituents might help us learn which spans
could be arguments.
Toward this end we use additional annotations from a separate
training corpus that does not significantly overlap the FrameNet corpus: the
Penn Treebank (PTB).\footnote{There is no overlap between the full-text test data
and PTB, however 86 sentences in the training and development data are also found in
PTB.}
Our multi-task learning setting allows us to learn span embeddings that are
shared between our frame-semantic parser and a model for predicting syntactic
constituents.

Formally, we consider a segment $\tuple{i,j,r}$ where $r \in \{0,1\}$ is a label with a corpus-specific definition. 
Spans in PTB starting at position $i$ and ending at position $j$ which are gold
constituents get $r^\ast = 1$, and others get $r^\ast = 0$.
Similarly, for FrameNet, we assign $r^\ast = 1$ for every span in a sentence
which has been annotated as a frame element for any frame, and $r^\ast = 0$ otherwise.

Analogous to the scoring function under our basic model~\sect{sect:argid}, we define a new scoring function, $\psi$ for every segment $\tuple{i,j,r}$:
\begin{align}
\psi(\tuple{i,j,r},\seq{x}) =& \tilde{\tensor{w}_2} \cdot \relu \{\tilde{\tensor{W}_1} [\hidden^{\text{span}}_{i:j}; \tensor{v}_{r}] \},
\end{align}
where $\tilde{\tensor{W}_1}$ and $ \tilde{\tensor{w}_2}$ are parameters of the model, $\tensor{v}_r$ is a learned embedding for the label $r$, and $\hidden^{\text{span}}_{i:j}$ is the span embedding, reused from our basic model~\sect{sect:spanemb}.

Our scaffold loss function is essentially a binary logistic regression
loss for each text span:
$\text{loss}_{\text{scaffold}}(\tuple{i,j,r^\ast}, \seq{x})=$
\begin{align}
 - \log \frac{
   \exp \psi(\tuple{i,j,r^\ast}, \seq{x})
   }{
     \displaystyle\sum_{r=\{0,1\}}{
       \exp \psi(\tuple{i,j,r}, \seq{x})
     }
   }.
\end{align}
The joint multi-task loss for a single sentence is:
\begin{align}
\underbrace{\text{loss}(\seq{x}, \seq{s}^\ast)}_{\text{Eq.~\ref{eqn:softmax-margin-segrnn}}} +  \delta \displaystyle\sum_{\substack{1 \leqslant i \leqslant j \leqslant |\seq{x}| \\ j-i < D}} \text{loss}_{\text{scaffold}}(\tuple{i,j,r^\ast}, \seq{x}),
\end{align}
where $\delta < 1$ is a hyperparameter used to de-emphasize the scaffold task,
tuned on the development set.
The first term does not apply to sentences in PTB,
since we do not have frame-semantic annotations $\seq{s}^\ast$.

At prediction time (including test time), no syntactic prediction is necessary; the scaffold is removed.  

\section{Experiments}
\label{sect:experiments}

In this section, we provide details of the dataset and experimental setup for all four models:  the basic SegRNN~\sect{sect:argid}, the phrase-structure and dependency syntactic feature additions~\sect{sect:synfeats} and the syntactic scaffold~\sect{sect:scaffold}.

\subsection{Data}
Our dataset contains sentences\footnote{2,716 train sentences with 17,141 targets and 2,420 test sentences with 4,427 targets.} from the full-text portion of FrameNet release 1.5 (September 2010).
We use the same test set as~\citet{Das:11} to facilitate comparison with related work. 
We chose eight additional files at random as a held-out development set; the rest of the files in the full-text data are used for training.
The FrameNet full-text data occassionally contains multiple annotations for the same target.  We use only the first annotation for such examples, following \citet{Fitzgerald:15}.

We use SyntaxNet~\citep{Andor:16} for predicted part-of-speech tags and Universal dependencies, from a released pretrained model.\footnote{\url{https://github.com/tensorflow/models/tree/master/syntaxnet}}
For phrase-structure parses, we use the RNNG parser~\citep{Dyer:16}, trained on WSJ \S2-21.
We stochastically (with probability 0.1) replace words that only appear once in the training data with an \textsc{unk} token to acquire estimates for out-of-vocabulary words at test time.


For the syntactic scaffold, we used all 49,208 sentences from WSJ \S00--24 of Penn Treebank.

\begin{figure}
\centering
  \includegraphics[width=.95\columnwidth]{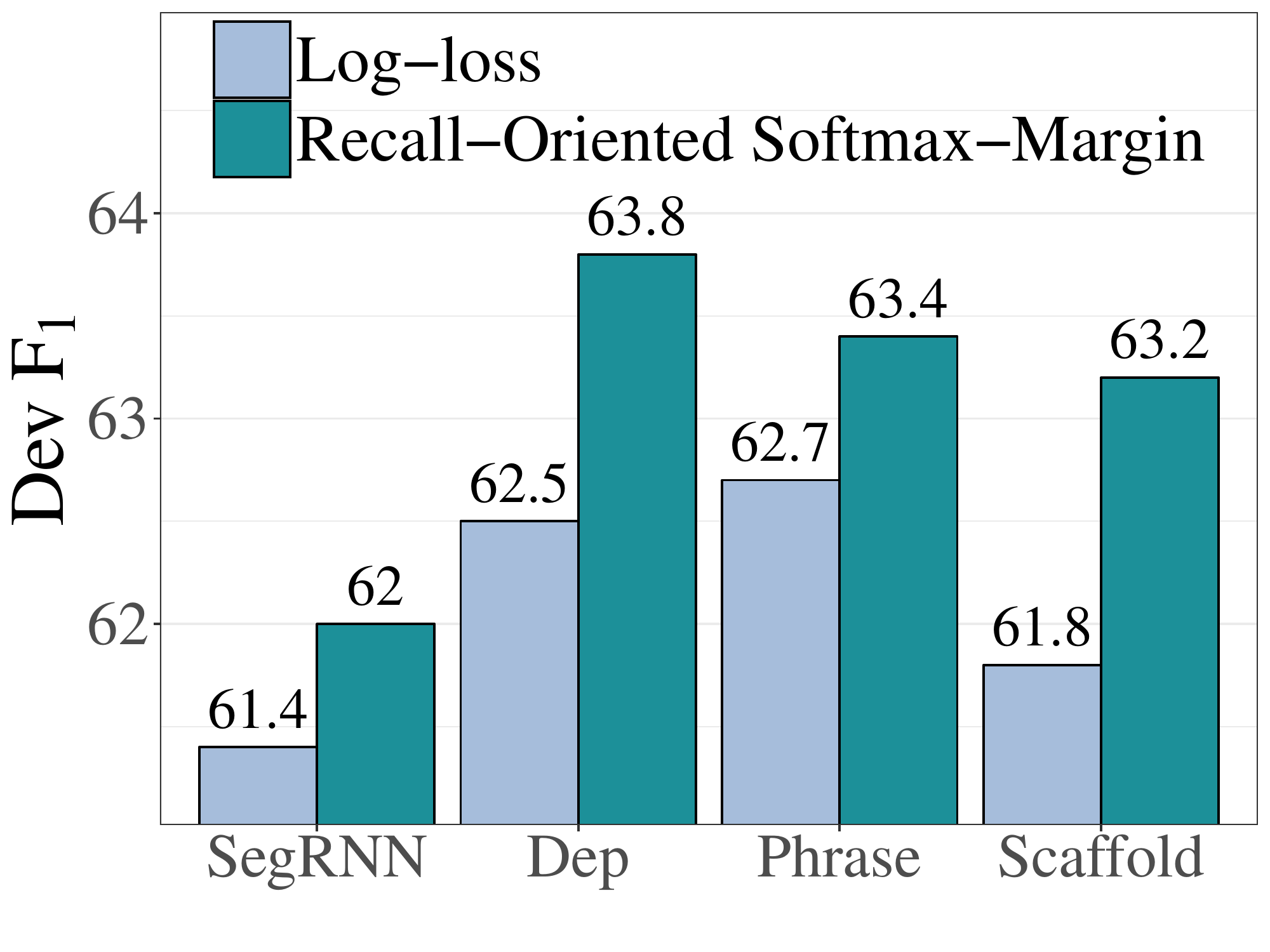}
  \caption{ Development-set $F_1$ with log-loss
    (no cost) vs.~recall-oriented cost.
  \label{fig:logloss}}
\end{figure}

\subsection{Hyperparameters}
\label{sect:hyperparams}

We used single-layer LSTMs for sentence encoding, spans, targets, dependency and nonterminal paths, each with a hidden state of size 64. 
Pretrained GloVe~\cite{Pennington:14} vectors of dimension 100 are used, trained on a corpus of 6 billion words; we do not update these during training.
Learned embeddings of size 60, 4, 100, 64, 50, 8 and 16 are used for words, POS tags, frames, lexical units, frame-elements, dependency labels and nonterminals, respectively.
For ADAM we set the initial learning rate to 0.0005,
the moving average parameter to 0.01,
the moving average variance to 0.9999,
and the $\epsilon$ parameter (to prevent numerical instability) to $10^{-8}$;
no learning rate decay is used.
To prevent ``exploding'' gradients, we clip the 2-norm of the gradient~\cite{Graves:13} to 5 before each gradient update.
These values were selected based on intuition and prior work;
a more careful tuning of the above hyperparameters could be expected to improve performance.

The remaining hyperparameters were chosen based on their $F_1$ performance on the held-out development set.
We selected the dropout rate~\cite{Srivastava:14} of 0.05 from the set $\{0.01,0.05,0.1\}$.
We selected the recall-oriented cost $\alpha=2$ from the set $\{1,2,5,10\}$.
We selected the scaffold weight $\delta=0.17$ from the set $\{0.17,0.34,0.89\}$.

Our experiments were run using the DyNet library~\cite{Neubig:17}.\footnote{\url{https://github.com/clab/dynet}}

\subsection{Self-Ensembling}
\label{sect:ensemble}

To compensate for the variance resulting from different initializations, we use a self-ensembling approach.
We train five models, differing only in their random initialization, and ensemble their local scores  at test time.
Specifically, we calculate the sum of the segment scores under each model ($\segScore(s,\seq{x})$ in Eq.~\ref{eqn:segScore}) to get the final ensembled segment score, which is then plugged into Eq.~\ref{eq:dp} to decode.

\subsection{Evaluation}
All systems are evaluated for precision, recall, and $F_1$, micro-averaged across test examples, following standard practice.
We use the standard script provided by SemEval 2007 \cite{Baker:07}, with a single modification provided by \citet{Kshirsagar:15} to optionally ignore the frame identification output.
This allows us to evaluate for argument identification in isolation, which is the primary focus of this paper; for this setting we use gold frames (without rewarding them in $F_1$ evaluation).
We also evaluate with predicted frames to illustrate our effect on end-to-end parsing performance;
for this, we use the same predicted frames as \citet{Fitzgerald:15}, who retrained the frame identification model from~\citet{Hermann:14} but with an updated dependency parser.

\begin{table*}[th]
    \footnotesize
    \centering
    \ra{1.1}
    \begin{tabular}{@{}l ccc c ccc@{}}
			\toprule
			& \multicolumn{3}{c}{Arg.~Id., Gold Frames}
			& \phantom{abc}
			& \multicolumn{3}{c}{Frame+Arg.~Id, Predicted Frames} \\
			\cmidrule{2-4}
			\cmidrule{6-8}
			& P & R & $F_1$ &
			& P & R & $F_1$ \\

			\midrule
			SEMAFOR (Das et al., 2014)
			& 65.6 & 53.8 & 59.1 &
			& - & - & 66.8\\

			$^\ast$SEMAFOR (Kshirsagar et al., 2015)
			& 66.0 & 60.4 & 63.1 &
			& - & - & 67.9 \\
			
			$^\ast$FitzGerald et al., 2015
			& - & - & - &
			& 74.8 & 65.5 & 69.9\\
			
			Roth et al., 2017
			& - & - & - &
			& 72.1 & 68.3 & 70.2\\

			$^\ast$FitzGerald et al., 2015 (ensemble)
			& - & - & - &
			& 75.0 & 67.3 & \textbf{70.9} \\
			
			\midrule[\cmidrulewidth]

			\sysname
			& 64.7 & 61.2 & 62.9 &
			& 68.0 & 68.1 & 68.0\\
			
			\sysname~+ dependency features
			& 69.4 & 60.5 & 64.6 &
			& 71.0 & 67.8 & 69.4 \\
			
			\sysname~+ phrase-structure features
			& 69.1 & 61.8 & 65.3 &
			& 70.4 & 68.3 & 69.3 \\
			
			\sysname~+ syntactic scaffolding
			& 68.4 & 60.7 & 64.4 &
			& 70.1 & 67.9 & 69.0\\

			\midrule[\cmidrulewidth]

			\sysname~(ensemble)
			& 69.5 & 63.6 & 66.4 &
			& 70.5 & 69.4 & 69.9\\
			
			\sysname~+ dependency features (ensemble)
			& 70.2 & 65.6 & 67.8 &
			& 70.7 & 70.4 & 70.6 \\
			
			\sysname~+ phrase-structure features (ensemble)
			& 71.7 & 66.3 & 68.9 &
			& 71.2 & 70.5 & \textbf{70.9} \\			

			\sysname~+ syntactic scaffolding (ensemble)
			& 72.0 & 65.0 & 68.3 &
			& 71.5 & 69.9 & 70.7 \\

		\bottomrule
		\end{tabular}
	\caption{Parsing results on the FrameNet 1.5 test set. The first
	three columns evaluate performance of argument identification only using gold frames.
	The last three columns are a combined evaluation of frame identification and argument identification
	together, using predicted frames from \citet{Fitzgerald:15}.
        $^\ast$These systems use additional semantic resources during training, a technique orthogonal to those presented in
        this paper.
        \sam{what does bolding indicate?}\sw{bolding indicates best
          numbers. Ideally we should have significance tests to
          confirm that scaffolding and dependency are also state of
          the art. Maybe in arXiv v2.} 
          \nascomment{add version numbers to Das and Kshirsagar SEMAFORs?  Are they 1 and 2?}
	}
	\label{tab:results}
\end{table*}

\subsection{Baselines}

SEMAFOR \citep{Das:14} is a widely used system that  
identifies frame-semantic arguments using a linear model with hand-engineered features based on dependency parses.
SEMAFOR also prunes out argument spans using syntactic heuristics and uses beam search, or optionally AD$^3$, to decode while respecting constraints \citep{Das:12}.
\citet{Kshirsagar:15} extended SEMAFOR through the use of exemplar FrameNet annotations, guide features from PropBank, and the FrameNet hierarchy.

In an extension to \citet{Tackstrom:15}, \citet{Fitzgerald:15} proposed a multi-task learning approach for frame-semantic parsing and PropBank SRL, using a feed-forward neural network to score candidate arguments.
The input to the neural network is a set of hand-engineered features extracted from a dependency parse. 

In a separate line of work, Framat~\cite{Roth:15} adds features based on context and discourse to improve an SRL system~\cite{Bjorkelund:10} adapted for frame-semantics, using a global model with reranking.
\citet{Roth:16} and \citet{Roth:17} extend this model by learning embeddings for dependency paths between targets and their arguments.
We borrow their use of path embeddings as syntactic features, but we explicitly model argument spans using SegRNNs, without any reranking.
More importantly, our scaffolding model does not rely on a syntactic parser.

\subsection{Results}

Table~\ref{tab:results} shows the performance of five published baseline systems, along with our four
new models (with and without ensembling).
Surprisingly, the basic model (\sysname) outperforms the original SEMAFOR
parser \citep{Das:14}, and matches an improved version
\citep{Kshirsagar:15}, without using any cues from syntax.
We attribute this improvement to representation learning.

Perhaps less surprisingly, all three syntactic additions improve the performance of our basic model.
The performances of phrase-structure features and dependency features are comparable.
The syntactic scaffold performance is not far behind, and in fact
beats the dependency features model after ensembling.
\sam{would be nice to say something about statistical significance?} \sw{Agreed, didn't get time for this :( Maybe in arXiv v2.}

Self-ensembling markedly improves performance across all models,
presumably because it reduces variance due to the non-convexity of the
learning objective.
Our best ensembled performance is tied with the state-of-the-art system by \citet{Fitzgerald:15}, which uses external semantic resources such as PropBank in addition to extensive syntactic features and self-ensembling.
Our best scaffolding model is within 0.2\% absolute $F_1$ of state-of-the-art, without any use of a syntactic parser or external semantic resources.


We also tested our recall-oriented softmax-margin loss
(Eq.~\ref{eqn:softmax-margin-segrnn}) against a plain log
loss.
Softmax-margin proved a good choice consistently across all our models
(see Fig.~\ref{fig:logloss}).

\section{Related Work}
\label{sec:related-work}

\paragraph{Joint syntax and semantics.} There has been much research on jointly modeling syntax and semantic roles, mostly on PropBank dependencies.
Other work has used parallel annotations to jointly model syntax and semantics, as in the CoNLL 2008--9 shared tasks \citep{Surdeanu:08,Hajivc:09}.
Such methods are able to more directly model the connection between syntax and semantics, but they require syntactic and semantic annotations over the same corpora \interalia{Johansson:08,Lluis:08,Titov:09,Roth:14,Lewis:15,Swayamdipta:16}.
Our work also learns syntactic representations, but needs only partial syntax annotations (bracketing), and not necessarily on the same data as that annotated with frame semantics.

\paragraph{Latent syntax} Some work has treated syntax as a latent variable and marginalized it out \citep{Naradowsky:12,Gormley:14}, an approach that requires no syntactic supervision, even during training, making it especially suitable for low-resource settings.
\citet{Collobert:11} use only syntactic boundaries for semantic role labeling; \citet{Zhou:15} extend their approach by forgoing all syntax and using very deep neural nets.
\citet{Kim:17} and \citet{Parikh:16} use different kinds of attention mechanism~\cite{Bahdanau:14} to learn representations of sentences for natural language inference.
To our knowledge, our basic model~\sect{sect:argid} is the first non-syntactic frame-semantic parser.

\section{Conclusion}
\label{sec:conclusion}

We presented a softmax-margin semi-Markov model that uses representation learning to predict frame-semantic arguments.
Our basic model achieves strong results without using any syntax.
We add syntax through a traditional pipeline as well as a multi-task learning approach which uses a syntactic scaffold only at training time.
Both approaches improve over the baseline, and achieve state-of-the-art performance, showing that syntax continues to be beneficial in frame-semantic parsing.
We conclude that scaffolding is a cheaper alternative to syntactic features since it does not require syntactic parsing at train or at test time.
Applying this technique to other tasks which rely on pipelining syntactic parsing is a promising avenue for future work.
Our parser is open-source and available at: 
\url{https://github.com/Noahs-ARK/open-sesame/}.

\section*{Acknowledgments}
We thank Adhiguna Kuncoro for help with RNNG, and Dipanjan Das and Michael Roth
for providing their output and for their helpful communication.
We also thank Hao Peng, George Mulcaire, Trang Tran, Kenton Lee, Luke
Zettlemoyer, and ARK members for valuable feedback.
This work was supported by DARPA grant FA8750-12-2-0342 funded under the DEFT program.

\bibliographystyle{emnlp_natbib}
\bibliography{framenet-emnlp17}


\newpage

\appendix
\section{Frame Identification}
\label{sect:frameid}

To complement our argument identification system, we also release a syntax-free frame identification system, which we describe here.
We treat frame id as an independent multiclass classification task for each target.
Given a sentence, a target, and its corresponding lexical unit, the task is to identify the frame the target evokes.
\citet{Hermann:14} and \citet{Fitzgerald:15} use gold targets and lexical units; for a fair comparison we do the same.

\subsection{Model}

Formally, the input to our model is a vector $\seq{x}= \tuple{\seq{w}_d, \seq{w}_o, t, \ell}$, following notation from~\S\ref{sect:notation}.
As before, $\seq{w}_d$ is an $n$-word sentence, $\seq{w}_o$ is its predicted part-of-speech tag sequence, $t = \tuple{t_{\text{start}}, t_{\text{end}}}$ is a target span in the sentence, and $\ell$ is its lexical unit.
For each lexical unit $\ell$, FrameNet provides the set of frames
$\textsc{F}_{\ell}$ that it could possibly evoke.

At each word position $q$, we give as input to a token biLSTM (f-tok) an input vector
$\tensor{u}_q = [\tensor{d}_{q}; \tensor{e}_{q}; \tensor{o}_q]$, where $\tensor{d}_{q}$ is a learned embedding of the word type, $\tensor{e}_{q}$
is a fixed pre-trained embedding of the word type, and $\tensor{o}_{q}$ is a learned embedding of the part-of-speech tag.
This yields a hidden state vector $\hidden^{\text{f-tok}}_q$ at the $q$th
token, corresponding to a contextualized representation of the word:
\begin{align}
\hidden^{\text{f-tok}}_{q} =& \left[\bilstm^{\text{f-tok}}(\tensor{u}_1,
                          \ldots, \tensor{u}_n)\right]_q .
\end{align}
To represent the target in context, we use $\hidden^{\text{f-tok}}$ over the target
span $t$, as well as the neighboring context window of size 1, as input to a forward LSTM:
\begin{equation}
\tensor{u}_{t} =
  \lstm^{\text{f-tgt}}(
    \hidden^{\text{f-tok}}_{t_{\text{start}}-1},
    \ldots,
    \hidden^{\text{f-tok}}_{t_{\text{end}}+1}).
\end{equation}
In addition to the above, we learn an embedding $\tensor{u}_{\ell}$ for each lexical unit $\ell$.
A final layer with a rectified linear unit~\cite{Nair:10} is applied to
$[\tensor{u}_{t}; \tensor{u}_{\ell}]$ to get our model scores:
\begin{align}
\label{eqn:frameScore}
  \nu(f) =&
    \ w^f_3 \ \relu \{ \tensor{w}^f_4 [\tensor{u}_{t}; \tensor{u}_{\ell}] \},
\end{align}
where the scalar $w^f_3$ and the vector $\tensor{w}^f_4$ are model parameters associated with the frame $f$.
The probability under our model is given by:
\begin{align}
\label{eqn:frameProb}
  p(f) =&
    \frac{
      \exp{\nu(f)}
    }{
      \displaystyle\sum_{f' \in \textsc{F}_{\ell}}{ \exp{\nu(f')}}
    }.
\end{align}
Negative log likelihood is minimized during training, using ADAM~\citep{Kingma:14}.
A single pass through the sentence is sufficient to predict frames for all targets.

%
%

To compensate for the variance resulting from different initializations, we use a five model self-ensemble (as in~\S\ref{sect:ensemble}) to aggregate frame scores in Eq.~\ref{eqn:frameScore}.
Pretrained GloVe~\cite{Pennington:14} vectors of dimension 100 are used.
For tuning hyperparameters, we use the same methods as described in~\S\ref{sect:hyperparams}. 

\subsection{Results}

The performance of our model is shown in Table~\ref{tab:frameidresults}.
Our model significantly outperforms SEMAFOR, which uses a variety of syntactic features.
\citet{Hermann:14} reuse most of the features from SEMAFOR, as well as WSABIE
embeddings~\cite{Weston:11} that also make use of syntax.
\citet{Fitzgerald:15} reimplement their model and report slightly better scores (possibly due to differences in parameter initialization or syntactic parses used).
Our performance is similar to \citet{Hartmann:17}, who focus on improving out of domain frame identification performance using a large external database.
We use no external resources.

\begin{table}
  \footnotesize
  \centering
  \begin{tabular}{@{}l c@{}}
    \toprule

    & Exact Match \\

    \midrule

    SEMAFOR~\cite{Das:14}              & 83.60 \\
    \citet{Hartmann:17}                & 87.63 \\
    \citet{Hermann:14} /               & \multirow{ 2}{*}{\textbf{88.41}}\\
    \hspace{.5cm}\citet{Fitzgerald:15} &  \\

    \sysname                           & 86.94 \\
    \sysname\ (ensemble)                & 87.51 \\

    \bottomrule

  \end{tabular}
  \caption{Test set accuracy of frame identification.}
  \label{tab:frameidresults}
\end{table}

Finally, we show the performance of our end-to-end system with both frame and
argument identification in Table~\ref{tab:fullfsp}.
As expected, it is lower than the results in Table~\ref{tab:results},
illustrating the importance of frame identification in a frame-semantic parsing
pipeline.

\begin{table}
  \footnotesize
  \centering
  \begin{tabular}{@{}l ccc@{}}
    \toprule

    & P & R & $F_1$ \\

    \midrule

    \sysname                              & 69.6 & 69.1 & 69.4 \\
    ~+ dependency features        & 70.0 & 70.1 & 70.0 \\
    ~+ phrase-structure features  & 70.6 & 70.4 & 70.5 \\
    ~+ syntactic scaffolding      & 69.9 & 70.9 & 70.4 \\

    \bottomrule
  \end{tabular}
  \caption{Full end-to-end performance, when combined with our ensembled arg id models.}
  \label{tab:fullfsp}
\end{table}

\end{document}